\documentclass[conference]{IEEEtran}
\IEEEoverridecommandlockouts
\makeatletter
\def\endthebibliography{%
  \def\@noitemerr{\@latex@warning{Empty `thebibliography' environment}}%
  \endlist
}
\makeatother
\usepackage{cite}
\usepackage{amsmath,amssymb,amsfonts}
\usepackage{algorithmic}
\usepackage{graphicx}
\usepackage{textcomp}
\usepackage{xcolor}
\def\BibTeX{{\rm B\kern-.05em{\sc i\kern-.025em b}\kern-.08em
    T\kern-.1667em\lower.7ex\hbox{E}\kern-.125emX}}
\begin{document}

\title{Ethical Challenges of Using Artificial Intelligence in Judiciary}

\author{\IEEEauthorblockN{Angel Mary John\thanks{Personal use of this material is permitted.  Permission from IEEE must be obtained for all other uses, in any current or future media, including reprinting/republishing this material for advertising or promotional purposes, creating new collective works, for resale or redistribution to servers or lists, or reuse of any copyrighted component of this work in other works. Published article \cite{john2023ethical}}}
\IEEEauthorblockA{\textit{Dept. of Law} \\
\textit{Mar Gregorios College of Law}\\
Thiruvananthapuram, India\\
angel@mgcl.ac.in}
\and
\IEEEauthorblockN{Aiswarya M. U.}
\IEEEauthorblockA{\textit{Dept. of Law} \\
\textit{Mar Gregorios College of Law}\\
Thiruvananthapuram, India\\
aiswaryamu@mgcl.ac.in}
\and
\IEEEauthorblockN{Jerrin Thomas Panachakel}
\IEEEauthorblockA{\textit{Dept. of Electronics and Comm. Engg.} \\
\textit{College of Engineering, Trivandrum}\\
Thiruvananthapuram, India \\
jerrin.panachakel@cet.ac.in}
}
\maketitle

\begin{abstract}

Artificial intelligence (AI) has emerged as a ubiquitous concept in numerous domains, including the legal system. AI has the potential to revolutionize the functioning of the judiciary and the dispensation of justice. Incorporating AI into the legal system offers the prospect of enhancing decision-making for judges, lawyers, and legal professionals, while concurrently providing the public with more streamlined, efficient, and cost-effective services. The integration of AI into the legal landscape offers manifold benefits, encompassing tasks such as document review, legal research, contract analysis, case prediction, and decision-making. By automating laborious and error-prone procedures, AI has the capacity to alleviate the burden associated with these arduous tasks. Consequently, courts around the world have begun embracing AI technology as a means to enhance the administration of justice. However, alongside its potential advantages, the use of AI in the judiciary poses a range of ethical challenges. These ethical quandaries must be duly addressed to ensure the responsible and equitable deployment of AI systems. This article delineates the principal ethical challenges entailed in employing AI within the judiciary and provides recommendations to effectively address these issues.
\end{abstract}


\section{Introduction}
The field of Artificial Intelligence (AI) has a long and rich history, and there have been many different AI systems developed over the years. The earliest AI systems were developed in the 1950s and 1960s, and they were focused primarily on performing tasks that would normally require human intelligence, such as problem-solving, reasoning, and language processing.

One of the earliest and most well-known AI systems is the Logic Theorist, which was developed by Allen Newell and Simon Herbert in 1956 \cite{newell1956logic}. The Logic Theorist was a computer program designed to prove mathematical theorems using symbolic logic. Another early AI system was the General Problem Solver (GPS), developed by Newell et al. in 1957, which could solve a wide range of problems by searching through a set of possible solutions \cite{newell1959report}. Other early AI systems included the perceptron, developed by McCulloch and Pitts in 1943, which was a type of neural network that could learn to recognize patterns \cite{rosenblatt1957perceptron}; and ELIZA, developed by Joseph Weizenbaum in 1966, which was a natural language processing system that could simulate a conversation between a human and a computer \cite{weizenbaum1966eliza}. Since then, the field of AI has continued to develop rapidly, with many new and innovative AI systems being developed each year. 

An important aspect of AI systems, which is often overlooked is the ethics or moral philosophy of AI systems. Ethics deals with moral principles regarding what is morally good \& bad and right \& wrong \cite{velasquez2002business}. This article discusses the ethical aspects of using AI in judiciary. 

\section{Major AI Systems in Judiciary}

Similar to other domains such natural language processing (NLP) \cite{mnasri2019recent}, computer vision (CV) \cite{khan2021machine}, public health \cite{baclic2020artificial}, biomedical engineering \cite{panachakel2021can,panachakel2021decoding} and finance \cite{cao2020ai}, the domain of judiciary has witnessed the influence of AI in various sub-domains. The major AI systems used in the judiciaries across the globe are:

 \begin{itemize}
     \item Giustizia Predittiva: Giustizia Predittiva is an AI-powered platform developed by the Italian Ministry of Justice. It can be used to predict the outcome of a case based on analysis of past cases and legal precedents.
     \item LITI: LITI is an AI-powered legal research tool developed by the Italian Bar Association. It can be used to search legal databases, analyze legal documents, and provide insights into legal issues.
    \item eCourts Services: This is an integrated case management system that is being used by various courts across India. It provides a range of services, including online case filing, case status tracking, and document management.
    \item Case Information and Management System (CIMS): CIMS is an AI-powered case management system that is being used by the Supreme Court of India and several High Courts in India. It helps in the automation of case management processes, including case filing, document management, and case tracking.
    \item Integrated Decision Support System (IDSS): IDSS is an AI-powered system that provides judges with relevant legal information, including case laws and precedents, to help them make informed decisions. It is being used in the High Courts of Delhi, Punjab and Haryana, and Bombay in India.
    \item Artificial Intelligence Legal Information Analysis System (AILIAS): AILIAS is an AI-powered legal research tool that is being used by the Delhi High Court in India. It uses NLP and machine learning algorithms to analyze and summarize legal documents.
    \item SUPACE: The Supreme Court Portal for Assistance in Courts Efficiency (SUPACE) is an AI-powered platform developed by the Supreme Court of India to improve the efficiency of court proceedings. The platform uses advanced analytics and machine learning algorithms to provide judges with relevant legal information and assist them in decision-making.
    \item SUVAS: The Supreme Court Vidhik Anuvaad Software (SUVAS) is an AI-powered translation software developed by the Supreme Court of India to assist judges and lawyers in translating legal documents from English to regional languages and vice versa. The software uses advanced machine learning algorithms and natural language processing (NLP) techniques to provide accurate and efficient translations.
    \item COMPAS (Correctional Offender Management Profiling for Alternative Sanctions) is a risk assessment tool used by some courts in the USA to evaluate a defendant's likelihood of reoffending and make recommendations for bail or sentencing. The system uses a combination of data analysis and machine learning algorithms to generate a risk score for each defendant based on various factors, such as criminal history, age, and employment status.
\end{itemize}
\section{Ethical Challenges of using AI in Judiciary}
There principle ethical challenges in using AI in judiciary, which are:
\begin{enumerate}
    \item Bias and fairness
    \item Transparency and accountability
    \item Privacy and data protection
    \item Speech imagery-based BCI systems
\end{enumerate}

\subsection{Bias and Fairness}
AI systems are primarily dependent on the data they are trained on. Therefore any biases in the criminal justice system in the data will be reflected in the trained model.  This can lead to unfair or discriminatory outcomes for certain groups, such as minorities, and raise concerns around due process and equal protection under the law.

Comparing the bias of an AI system to human psychology, the bias can be termed as an implicit bias where both the AI system and its developers are unaware of the bias \cite{dovidio2002implicit,maryfield2018implicit}. Due to the lack of the awareness of the bias, the developers do not fine tune the AI model to mitigate this bias. For instance, consider an AI system trained to assist in law enforcement professionals in drug-related incidents. Since currently disproportionately large number of people of colour are prosecuted in drug-related cases \cite{maryfield2018implicit,tonry2010social}, an AI system trained on this data will also exhibit a racial bias similar to the implicit racial bias in humans \cite{dovidio2002implicit, bilotta2019subtle}. That is, the system is not explicitly trained to exhibit this bias but the data the system has seen during the training phase has made system to exhibit racial bias. A notable case of bias AI system used in law enforcement is COMPAS or Correctional Offender Management Profiling for Alternative Sanctions which was shown to be biased against African-Americans \cite{larsson2019socio}.

\subsection{Transparency and Accountability}
An emerging area in the domain of AI is explainable AI or XAI. XAI are those algorithms that consider the human comprehensibility as one of the considerations \cite{larssontransparency, larsson2019socio}. AI transparency is a related terminology which means making ``black-box'' AI algorithms more human understandable.  Accountability in AI refers to the principle and practice of holding AI systems, their developers, and users responsible for the behavior, decisions, and outcomes of the AI system \cite{floridi2018ai4people}. Transparency and accountability are two of the seven key requirements that AI systems should meet in order to be deemed trustworthy, according to the EU's  High-Level Expert Group on AI presented Ethics Guidelines for Trustworthy Artificial Intelligence.

AI systems can be opaque and difficult to understand, which can make it challenging to determine how they arrived at a particular decision. This lack of transparency can undermine public trust in the judiciary and raise concerns around accountability and due process. When an AI system is used to help a judicial officer in making a decision, it is important for the officer to know why the AI system is making a particular recommendation. This helps the officer to identify any implicit biases in the recommendations of the AI system. The judgments passed by a judicial officer in each case is highly dependent on the facts of the case. It is often possible that the judgement for two cases with seemingly similar facts can be entirely different due to the nuances in the facts. An AI system trained on large amount of data may miss these nuances, leading to incorrect recommendations. A transparent AI system enables human beings to inspect whether the AI system has considered these nuances and the influence of the input data on the decision. 

\subsection{Privacy and Data Protection}
Artificial intelligence (AI) systems heavily rely on vast amounts of data to train their algorithms and improve their performance. In the context of the criminal justice system, this data may encompass highly sensitive and personal information related to individuals involved in legal proceedings, such as defendants, victims, witnesses, and even law enforcement personnel.

The ethical and legal challenges arise from the potential mishandling or misuse of this data, which can lead to severe breaches of privacy and data protection regulations. The nature of the information contained within these datasets, including details about criminal records, health conditions, financial records, and other personally identifiable information, requires careful handling and safeguarding.

Mishandling data in AI systems can manifest in various ways. It could involve unauthorized access, storage in unsecured environments, inappropriate sharing or distribution, or even accidental or intentional manipulation of the data. Such lapses in data security not only violate privacy rights but can also undermine public trust in the criminal justice system and AI technologies as a whole.

The consequences of data breaches in AI systems used within the judiciary are multifaceted. They can result in the exposure of sensitive information to unauthorized individuals or entities, leading to identity theft, discrimination, reputational harm, or even physical harm in extreme cases. Moreover, data breaches can impede the fair administration of justice by compromising the integrity and confidentiality of legal proceedings.

\subsection{Speech Imagery-based BCI Systems}

Recently, researchers have developed brain–computer interface (BCI) systems for decoding imagined speech from various neuroimaging modalities \cite{hiremath2015brain, proix2022imagined, panachakel2021can, panachakel2021decoding}. Imagined speech is defined as speaking without moving the articulators \cite{panachakel2021decoding1}. These systems can be extended for decoding the thoughts of human beings, which can be used for eliciting confessions from the accused. Although there are legislation against forced confessions, the lack of specific guidelines in using speech-imagery based BCI systems in interrogation raises several ethical questions such as 
\begin{enumerate}
    \item Reliability and Accuracy: The reliability and accuracy of information obtained through speech-imagery based BCI systems are highly contentious. \item Violation of Human Rights: The use of speech-imagery based BCI systems may be perceived as a violation of an individual's right to protection against self-incrimination or the right to privacy.
    \item Legal and Judicial Reliability: The admissibility and reliability of evidence obtained through speech-imagery based BCI systems can be contentious in legal proceedings. 

\end{enumerate}
\section{Recommendations to address the ethical challenges of using AI in judiciary}
\subsection{Human Oversight and Responsibility}
AI systems should not replace human judgement and decision-making entirely, as this could lead to a lack of accountability and responsibility for decisions made. It is important to ensure that human oversight is maintained and that the ultimate responsibility for decisions made using AI remains with human judges and legal professionals. The following steps help to address the ethical challenges of using AI in judiciary by human oversight:
\begin{enumerate}
    \item Decision-making and Final Authority: Human oversight ensures that the ultimate decision-making authority rests with human judges or legal professionals. While AI systems can provide recommendations or insights, the final judgment and decision-making power remain in the hands of humans. This prevents the complete delegation of decision-making to AI and ensures that human values, context, and legal expertise are considered.
    \item Bias Detection and Mitigation: Human oversight can help in identifying and addressing biases in AI systems. Humans can review and evaluate the outputs of AI algorithms, detect any unfair biases or discriminatory patterns, and take corrective measures to mitigate them. Human judges can assess the fairness and legality of AI-generated decisions and intervene if necessary.
    \item Explainability and Transparency: Human judges and legal professionals can demand explanations from AI systems about their decision-making processes. When AI systems are not able to provide clear and understandable explanations, human oversight can question and challenge their outputs. This promotes transparency and accountability, enabling the identification of potential errors, biases, or unjust outcomes.
    \item Contextual Considerations: Human judges bring contextual knowledge, experience, and legal reasoning to the decision-making process. They can consider nuances, legal precedents, and specific circumstances that may not be adequately captured by AI systems. Human oversight ensures that decisions are made with a comprehensive understanding of the legal, social, and ethical implications of the case at hand.
    \item Ethical Oversight and Accountability: Human oversight enables ethical considerations to be continuously evaluated and updated as societal values evolve. Human judges and legal professionals can assess the ethical implications of AI systems and intervene if they perceive any conflicts with legal or ethical standards. They can also hold AI developers, operators, and users accountable for the ethical implications and consequences of the system's decisions.
    \item Adaptation to Changing Legal Landscape: Human oversight allows for adaptation to changing legal frameworks and regulations. AI systems in the judiciary need to align with legal requirements and comply with evolving ethical standards. Human judges can provide legal interpretations, ensure compliance with legal precedents, and adapt the AI system to changing legal and ethical landscapes.
\end{enumerate}

\subsection{Privacy and Data Protection}
The following steps can improve privacy and data protection:
\begin{enumerate}
    \item Data Minimization: AI systems should only collect and process the minimum amount of personal data necessary to achieve the intended purpose. Unnecessary data should be avoided to minimize privacy risks.
    \item Informed Consent: Individuals whose data is being processed should be provided with clear and understandable information about how their data will be used in AI systems. Consent should be obtained in a transparent and informed manner, ensuring individuals understand the implications of their data being used in the judicial AI system.
    \item Anonymization and Pseudonymization: Personal data used in AI systems should be appropriately anonymized or pseudonymized to reduce the risk of re-identification. This ensures that individuals cannot be directly identified from the data.
    \item Data Security: Strong security measures should be implemented to protect the confidentiality, integrity, and availability of the data used in the AI system. This includes encryption, access controls, secure storage, and regular security audits.
    \item Purpose Limitation: The personal data collected for the judiciary AI system should only be used for the specific purpose for which it was collected. Any secondary use of the data should be limited and clearly defined.
    \item Data Retention: Personal data should not be retained for longer than necessary. Clear guidelines should be established for data retention periods, and data should be securely and permanently deleted when it is no longer needed.
    \item Third-Party Data Sharing: If personal data is shared with third parties for the AI system, appropriate data protection agreements and safeguards should be in place to ensure the data is handled in compliance with privacy regulations.
    \item Accountability and Compliance: The judiciary AI system's developers and operators should be accountable for ensuring compliance with relevant privacy laws and regulations. Regular audits and assessments should be conducted to monitor and verify compliance.
\end{enumerate}
\subsection{Collaboration and Stakeholder Engagement}
Fostering collaboration among stakeholders is crucial for addressing ethical challenges in the use of AI in the judiciary. Collaboration among different stakeholders can help in:

\begin{enumerate}
    \item Exploring Diverse Perspectives: Collaboration brings together stakeholders with diverse backgrounds and expertise. Judges, legal professionals, AI developers, ethicists, and civil society organizations each bring unique perspectives and insights to the table. By engaging in collaborative discussions, these stakeholders can contribute their knowledge and viewpoints, helping to identify and address ethical challenges from multiple angles.
    \item Developing Comprehensive Understanding: Collaboration allows for a comprehensive understanding of the ethical challenges associated with AI in the judiciary. Each stakeholder group can contribute their specific expertise and domain knowledge to analyze the ethical implications of AI systems. This holistic understanding enables the development of more effective and well-informed solutions.
    \item Identifying Blind Spots: Different stakeholders may have different blind spots when it comes to ethical considerations. By collaborating, these blind spots can be identified and addressed. For example, AI developers may have technical expertise but may not fully understand the legal and ethical nuances, while legal professionals may have legal expertise but may not be aware of the technical limitations and potential biases in AI algorithms. Collaboration bridges these gaps and ensures a more comprehensive approach to addressing ethical challenges.
    \item Sharing Best Practices: Collaboration allows for the sharing of best practices and lessons learned from different contexts. Stakeholders can share their experiences, case studies, and insights about ethical challenges and how they have been addressed. This exchange of knowledge helps in identifying effective strategies and solutions that can be applied in different jurisdictions or scenarios.
    \item Co-creation of Solutions: Collaboration facilitates the co-creation of solutions that consider the needs and concerns of different stakeholders. By involving judges, legal professionals, AI developers, ethicists, and civil society organizations in the decision-making process, the resulting solutions are more likely to be well-rounded, balanced, and acceptable to all parties. It ensures that ethical considerations are integrated from the outset, leading to more responsible and accountable AI systems.
    \item Improving Public Trust and Legitimacy: Collaboration with diverse stakeholders helps in building public trust and legitimacy in the use of AI in the judiciary. By involving civil society organizations and engaging in transparent dialogues, the concerns and perspectives of the broader public can be taken into account. This inclusivity fosters trust, as it demonstrates that the ethical challenges are being addressed collectively, with the interests of all stakeholders in mind.
    \item Enabling Continuous Learning and Adaptation: Collaboration allows for ongoing learning and adaptation as new ethical challenges emerge or as societal values evolve. By maintaining an ongoing dialogue among stakeholders, they can stay informed about the latest developments, research findings, and regulatory updates in the field of AI ethics. This ensures that ethical frameworks and practices remain up to date and responsive to changing needs.
\end{enumerate}
\subsection{Potential for Errors and Unintended Consequences}
Like any technology, AI systems are not infallible, and there is a risk of errors and unintended consequences. This is particularly concerning in the context of the criminal justice system, where even small errors or biases could have significant impacts on people's lives and liberties. Using algorithms such as SHAP (SHapley Additive exPlanations) \cite{lundberg2021game} developed for explaining the outputs of AI systems can help in identifying errors including biases in an AI system . 
\subsection{Ethical Guidelines and Principles}
Developing a set of ethical guidelines and principles that should govern the use of AI in the judiciary will help in addressing the ethical challenges of using AI in judiciary in the followiong ways:
\begin{enumerate}
    \item Framework for Ethical Decision-making: Ethical guidelines provide a framework for making ethical decisions when designing, deploying, and using AI systems. They establish principles, values, and norms that guide the development and operation of AI systems. By following these guidelines, developers and users can ensure that ethical considerations are taken into account throughout the entire lifecycle of the AI system.
    \item Awareness of Ethical Issues: Ethical guidelines raise awareness about the potential ethical challenges and dilemmas that can arise in the context of AI systems. They help stakeholders, including AI developers, policymakers, and users, understand the ethical implications of AI technologies. By explicitly stating the ethical concerns, guidelines ensure that these issues are not overlooked or dismissed in the development and deployment processes.
    \item Guidance for Responsible Practices: Ethical guidelines offer specific recommendations and practices to promote responsible AI development and use. They provide guidance on topics such as fairness, transparency, accountability, privacy, bias detection, and mitigation. Following these guidelines helps developers and users adopt responsible practices, minimizing potential ethical risks and negative impacts associated with AI systems.
    \item Consistency and Uniformity: Ethical guidelines promote consistency and uniformity in ethical standards across different AI systems and applications. They help establish a common understanding of what is considered ethical and acceptable in the development and use of AI. This consistency ensures that ethical challenges are addressed consistently across different AI systems, enhancing fairness, transparency, and accountability.
    \item Stakeholder Engagement and Alignment: Developing ethical guidelines often involves engaging with various stakeholders, including experts, researchers, policymakers, industry representatives, and civil society organizations. This stakeholder engagement process helps align different perspectives and interests, fostering a collective agreement on ethical standards. By involving diverse stakeholders, ethical guidelines can incorporate a wide range of viewpoints and ensure that the interests of different parties are considered.
    \item Regulatory Compliance: Ethical guidelines can serve as a reference point for regulatory frameworks and legal requirements related to AI systems. They provide a basis for developing and enforcing regulations that address ethical concerns. Compliance with these guidelines can help organizations and individuals demonstrate that they are adhering to ethical principles, facilitating regulatory compliance and reducing the risk of legal or reputational consequences.
    \item Continuous Improvement and Adaptation: Ethical guidelines are not static but evolve over time to keep pace with technological advancements and changing ethical considerations. They provide a foundation for ongoing reflection, evaluation, and improvement of AI systems. As new ethical challenges emerge, guidelines can be updated to address them, ensuring that AI systems remain aligned with evolving ethical standards.
\end{enumerate}

These guidelines should align with legal requirements, respect human rights, and address the identified ethical challenges. Consider relevant principles such as fairness, transparency, accountability, human oversight, privacy, and non-discrimination. Ethics guidelines for trustworthy AI by EU's High-Level Expert Group on AI presented Ethics Guidelines for Trustworthy Artificial Intelligence is a noteworthy ethical guideline for general AI systems. 

\section{Conclusion}

The potential benefits inherent in the utilization of artificial intelligence (AI) systems within the judiciary are indeed considerable, promising a paradigm shift in the way legal processes unfold. With AI at their disposal, judges and legal professionals can leverage its capabilities to swiftly process vast quantities of data, ensuring efficient and accurate analysis. The ability of AI to detect intricate patterns and trends that might elude human perception holds immense potential for enhancing the accuracy and effectiveness of judicial decision-making. By integrating AI tools into their workflow, legal professionals can access comprehensive and nuanced insights that aid in forming well-informed and consistent judgments.

However, the integration of AI systems in the judiciary necessitates a conscientious approach, as it inevitably poses significant ethical challenges. These challenges must be addressed comprehensively to ensure that the implementation of AI in the legal domain adheres to responsible practices and upholds the principles of justice and fairness.

Ethical considerations arise from the immense power and potential impact of AI systems on legal proceedings. Questions surrounding transparency, accountability, and bias become critical focal points. It is imperative to establish transparent mechanisms that elucidate the decision-making processes of AI systems, ensuring that their outcomes can be understood and scrutinized by all relevant stakeholders. Additionally, holding developers and users accountable for the behavior and decisions of AI systems is crucial, as it ensures that the benefits and potential risks associated with AI technology are properly managed.

Bias, whether implicit or explicit, is another ethical challenge that must be addressed. AI systems can inadvertently perpetuate biases present in the data they are trained on, resulting in discriminatory outcomes. It is essential to continuously monitor and mitigate bias in AI algorithms, striving for fairness and equal treatment within the legal system.

Furthermore, the ethical implications of AI in relation to privacy and data protection cannot be overlooked. AI systems rely on vast amounts of data, some of which may be sensitive or personal in nature. Safeguarding individual privacy rights and ensuring the responsible use and storage of data becomes paramount to maintain public trust and confidence in the judiciary.

To effectively address these ethical challenges, a multidisciplinary approach is necessary. Collaboration among judges, legal professionals, AI developers, ethicists, and civil society organizations can facilitate the development of ethical guidelines and best practices. By engaging in open and inclusive discussions, stakeholders can collectively navigate the complex ethical landscape and establish frameworks that safeguard fundamental rights and values.
\bibliography{references.bib}

\begin{thebibliography}{10}
\providecommand{\url}[1]{#1}
\csname url@samestyle\endcsname
\providecommand{\newblock}{\relax}
\providecommand{\bibinfo}[2]{#2}
\providecommand{\BIBentrySTDinterwordspacing}{\spaceskip=0pt\relax}
\providecommand{\BIBentryALTinterwordstretchfactor}{4}
\providecommand{\BIBentryALTinterwordspacing}{\spaceskip=\fontdimen2\font plus
\BIBentryALTinterwordstretchfactor\fontdimen3\font minus
  \fontdimen4\font\relax}
\providecommand{\BIBforeignlanguage}[2]{{%
\expandafter\ifx\csname l@#1\endcsname\relax
\typeout{** WARNING: IEEEtran.bst: No hyphenation pattern has been}%
\typeout{** loaded for the language `#1'. Using the pattern for}%
\typeout{** the default language instead.}%
\else
\language=\csname l@#1\endcsname
\fi
#2}}
\providecommand{\BIBdecl}{\relax}
\BIBdecl

\bibitem{john2023ethical}
A.~M. John, M.~Aiswarya, and J.~T. Panachakel, ``Ethical challenges of using
  artificial intelligence in judiciary,'' in \emph{2023 IEEE International
  Conference on Metrology for eXtended Reality, Artificial Intelligence and
  Neural Engineering (MetroXRAINE)}.\hskip 1em plus 0.5em minus 0.4em\relax
  IEEE, 2023, pp. 723--728.

\bibitem{newell1956logic}
A.~Newell and H.~Simon, ``The logic theory machine-a complex information
  processing system,'' \emph{IRE Transactions on information theory}, vol.~2,
  no.~3, pp. 61--79, 1956.

\bibitem{newell1959report}
A.~Newell, J.~C. Shaw, and H.~A. Simon, ``Report on a general problem solving
  program,'' in \emph{IFIP congress}, vol. 256.\hskip 1em plus 0.5em minus
  0.4em\relax Pittsburgh, PA, 1959, p.~64.

\bibitem{rosenblatt1957perceptron}
F.~Rosenblatt, \emph{The perceptron, a perceiving and recognizing automaton
  Project Para}.\hskip 1em plus 0.5em minus 0.4em\relax Cornell Aeronautical
  Laboratory, 1957.

\bibitem{weizenbaum1966eliza}
J.~Weizenbaum, ``Eliza—a computer program for the study of natural language
  communication between man and machine,'' \emph{Communications of the ACM},
  vol.~9, no.~1, pp. 36--45, 1966.

\bibitem{velasquez2002business}
M.~G. Velasquez and M.~Velazquez, \emph{Business ethics: Concepts and
  cases}.\hskip 1em plus 0.5em minus 0.4em\relax Prentice Hall Upper Saddle
  River, NJ, 2002, vol. 111.

\bibitem{mnasri2019recent}
M.~Mnasri, ``Recent advances in conversational nlp: Towards the standardization
  of chatbot building,'' \emph{arXiv preprint arXiv:1903.09025}, 2019.

\bibitem{khan2021machine}
A.~A. Khan, A.~A. Laghari, and S.~A. Awan, ``Machine learning in computer
  vision: a review,'' \emph{EAI Endorsed Transactions on Scalable Information
  Systems}, vol.~8, no.~32, pp. e4--e4, 2021.

\bibitem{baclic2020artificial}
O.~Baclic, M.~Tunis, K.~Young, C.~Doan, H.~Swerdfeger, and J.~Schonfeld,
  ``Artificial intelligence in public health: Challenges and opportunities for
  public health made possible by advances in natural language processing,''
  \emph{Canada Communicable Disease Report}, vol.~46, no.~6, p. 161, 2020.

\bibitem{panachakel2021can}
J.~T. Panachakel, K.~Sharma, A.~Anusha, and A.~Ramakrishnan, ``Can we identify
  the category of imagined phoneme from {EEG}?'' in \emph{2021 43rd Annual
  International Conference of the IEEE Engineering in Medicine \& Biology
  Society (EMBC)}.\hskip 1em plus 0.5em minus 0.4em\relax IEEE, 2021, pp.
  459--462.

\bibitem{panachakel2021decoding}
J.~T. Panachakel and R.~A. Ganesan, ``Decoding imagined speech from {EEG} using
  transfer learning,'' \emph{IEEE Access}, vol.~9, pp. 135\,371--135\,383,
  2021.

\bibitem{cao2020ai}
L.~Cao, ``{AI} in finance: A review,'' \emph{Available at SSRN 3647625}, 2020.

\bibitem{dovidio2002implicit}
J.~F. Dovidio, K.~Kawakami, and S.~L. Gaertner, ``Implicit and explicit
  prejudice and interracial interaction.'' \emph{Journal of personality and
  social psychology}, vol.~82, no.~1, p.~62, 2002.

\bibitem{maryfield2018implicit}
B.~Maryfield, ``Implicit racial bias,'' \emph{Justice Research and Statistics
  Association}, 2018.

\bibitem{tonry2010social}
M.~Tonry, ``The social, psychological, and political causes of racial
  disparities in the american criminal justice system,'' \emph{Crime and
  justice}, vol.~39, no.~1, pp. 273--312, 2010.

\bibitem{bilotta2019subtle}
I.~Bilotta, A.~Corrington, S.~A. Mendoza, I.~Watson, and E.~King, ``How subtle
  bias infects the law,'' \emph{Annual review of law and social science},
  vol.~15, pp. 227--245, 2019.

\bibitem{larsson2019socio}
S.~Larsson, ``The socio-legal relevance of artificial intelligence,''
  \emph{Droit et soci{\'e}t{\'e}}, vol. 103, no.~3, pp. 573--593, 2019.

\bibitem{larssontransparency}
S.~Larsson and F.~Heintz, ``Transparency in artificial intelligence. internet
  policy rev. 9 (2), 1--16 (2020).''

\bibitem{floridi2018ai4people}
L.~Floridi, J.~Cowls, M.~Beltrametti, R.~Chatila, P.~Chazerand, V.~Dignum,
  C.~Luetge, R.~Madelin, U.~Pagallo, F.~Rossi \emph{et~al.}, ``Ai4people—an
  ethical framework for a good ai society: opportunities, risks, principles,
  and recommendations,'' \emph{Minds and machines}, vol.~28, pp. 689--707,
  2018.

\bibitem{hiremath2015brain}
S.~V. Hiremath, W.~Chen, W.~Wang, S.~Foldes, Y.~Yang, E.~C. Tyler-Kabara, J.~L.
  Collinger, and M.~L. Boninger, ``Brain computer interface learning for
  systems based on electrocorticography and intracortical microelectrode
  arrays,'' \emph{Frontiers in integrative neuroscience}, vol.~9, p.~40, 2015.

\bibitem{proix2022imagined}
T.~Proix, J.~Delgado~Saa, A.~Christen, S.~Martin, B.~N. Pasley, R.~T. Knight,
  X.~Tian, D.~Poeppel, W.~K. Doyle, O.~Devinsky \emph{et~al.}, ``Imagined
  speech can be decoded from low-and cross-frequency intracranial {EEG}
  features,'' \emph{Nature communications}, vol.~13, no.~1, p.~48, 2022.

\bibitem{panachakel2021decoding1}
J.~T. Panachakel and A.~G. Ramakrishnan, ``Decoding covert speech from {EEG}-a
  comprehensive review,'' \emph{Frontiers in Neuroscience}, vol.~15, p. 392,
  2021.

\bibitem{lundberg2021game}
S.~Lundberg and S.~Lee, ``A game theoretic approach to explain the output of
  any machine learning model,'' \emph{Github: San Francisco, CA, USA}, 2021.

\end{thebibliography}
\bibliographystyle{IEEEtran}

\end{document}